\documentclass[]{article}
\usepackage[letterpaper]{geometry}
\usepackage{mtsummit2017}
\usepackage{times}
\usepackage{url}
\usepackage{latexsym}
\usepackage{natbib}
\usepackage{layout}

\usepackage{graphicx}

\begin{document}


\title{Neural and Statistical Methods for Leveraging Meta-information in Machine Translation}

\author{\name{\bf Shahram Khadivi} \hfill  \addr{skhadivi@ebay.com}\\
\addr{eBay Inc., Aachen, Germany}
\AND        
		\name{\bf Patrick Wilken}\thanks{~~Patrick Wilken has contributed to this work during his internship at eBay Inc.} \hfill 		\addr{patrick.wilken@rwth-aachen.de}\\   
        \addr{RWTH Aachen University, Aachen, Germany}
\AND        
        \name{\bf Leonard Dahlmann} \hfill \addr{fdahlmann@ebay.com}\\
        \name{\bf Evgeny Matusov} \hfill \addr{ematusov@ebay.com}\\
        \addr{eBay Inc., Aachen, Germany}
}


\maketitle
\pagestyle{empty}

\begin{abstract}
In this paper, we discuss different methods which use meta information and richer context that may accompany source language input to improve machine translation quality. We focus on category information of input text as meta information, but the proposed methods can be extended to all textual and non-textual meta information that might be available for the input text or automatically predicted using the text content. 
The main novelty of this work is to use state-of-the-art neural network methods to tackle this problem within a statistical machine translation (SMT) framework. 
We observe translation quality improvements up to 3\% in terms of BLEU score in some text categories.
\end{abstract}

\section{Introduction}
Using larger context in machine translation to improve its quality, including selection of correct word meaning, has been a challenging task.
Correct translation of polysemous words is vital to transfer important information from source sentence to the translation. To find the correct sense of polysemous words and phrases, usually only the context of the source sentence is available. Depending on the use case, the context can be extended to the surrounding sentences, or external signals about the text, like its topic or genre. Therefore, we can consider all methods that try to use a larger context for translation as methods that can help MT system select the right translation for polysemous words. In e-commerce, the problem of polysemous words is more severe. For example, incorrect literal translation of a brand name like Apple, Coach, Diesel, Affliction, Avenue, Cables To Go, Free People, etc. misleads a potential buyer. It also may create legal issues when e.g. a wrong translation directs buyers to a competitive brand name. In e-commerce MT scenarios, one of the main tasks is often the translation of the titles of the items offered for sale. Such titles are short, non-grammatical, and the local context of a given word is very variational. Since item title translation has a crucial importance in cross-border trade for e-commerce, we are trying to leverage meta data available for each item to deal with these irregularities.
Items in e-commerce inventory are usually classified according to a hierarchical taxonomy. The hierarchy itself contains top-level categories (like \textit{Clothing}, \textit{Electronics}) with varying degrees of depth in each top category. 
Although such hierarchy is created based on business insights, it implicitly groups objects which can be described in semantically similar terms. Therefore, we expect less variation in 
word senses within a category, with ambiguity decreasing deeper in the tree. For instance, Apple is very likely to be a brand name, not a fruit in \textit{Smartphones}. Therefore,  item category information can potentially provide a strong signal to identify the meaning of a word. 
In this work, we focus on using more broad product categories of a particular e-commerce site, on the top level 1 ($L_1$, e.g. ``Clothing and Shoes''), but also on levels 2 ($L_2$, e.g. ``Women's Shoes'') and 3 ($L_3$, e.g. ``Women's Boots''). 

The main goal of this work is to modify the major state-of-the-art approaches that leverage larger context in MT to use (category) meta-information both in training and at runtime and check experimentally whether such approaches are able to better translate polysemous words in e-commerce data and improve MT quality both overall and in specific categories. Among others, we look at neural machine translation (NMT) models which are by now state-of-the-art and by definition use larger context during decoding, as in \citep{bahdanau2014neural}. 
Since our goal is to enhance a real-time production phrase-based SMT system for title translation, and also to have a better understanding of the power of NMT as compared to SMT in using larger context, we use a bidirectional recurrent neural network (RNN) lexical model and also a feed-forward NN model as features in our SMT system. To the best of our knowledge, we were the first to modify these specific NMT models to use the embedding of sentence meta-information as an additional signal. Also, in this work we propose and test a simple generative category-specific word lexicon model.

The main challenges we encountered are data sparseness, data bias towards most frequently observed meaning of polysemous words, and absence of meta information for parts of training data. Nevertheless, significant improvements of MT quality could be achieved with some of the described methods on selected tasks.

In the next section, we describe major research works that employ larger context in MT. In Section~\ref{sec:models}, we describe the models we investigate in our work and the novel ways of integrating the category information into these models. Section~\ref{sec:experiments} is devoted to experimental results which include automatic and human evaluation on an English-to-Italian e-commerce title translation task. The paper concludes with a summary and future works in Section~\ref{sec:conclusion}. 

\section{Related Work}\label{sec:relwork}

Phrase-based MT models have an intrinsic problem in using large(r) context and long range dependencies, 
since these models are bounded by phrase context. Therefore, many research publications target solving these well-known problems of phrase-based MT. Here, we focus on pioneering works that are most comparable to this work. One research line of using larger context in MT is to use sentence context for word sense disambiguation ~\citep{Carpuat:emnlp2007}. \citet{Mauser:emnlp09} proposed the idea of employing a  discriminative lexical model that uses sentence level information to predict a target word, this idea has been extended and enhanced in a recent work \citep{Tamchyna:acl6}. \cite{Tamchyna:acl6} have proposed to extend the discriminative model to also use the target prefix to predict the next target word, and also they enhance the model to calculate target word probabilities on-line during the search using a fast and efficient classification method based on the~\texttt{Vowpal Wabbit\footnote{http://hunch.net/\textasciitilde{}vw/}} toolkit. \citet{Devlin:acl14} proposed a neural joint lexical model that also employs a larger context around a given source word including previous generated target words, to predict the corresponding target word probability using a feed-forward neural network as the classifier.

Research works of \citet{OSM-2011-Durrani} and \citep{Guta:JTR:wmt15} can also be considered as attempts to use larger context and long-range dependencies in phrase-based MT by modeling the dependencies between phrases.

The use of topic models in MT is another promising way to use larger context in lexical translation. The main idea is to use the topic models to infer the topic of the whole document/sentence, and then use it as a signal to the MT system to find the correct sense of the source word to translate~\citep{ACL-2012-Eidelman}. Hasler et al. have investigated different ways to use topic models in SMT~\citep{IWSLT-2012-Hasler,AMTA-2014-Hasler,EACL-2014-Hasler}. They showed relatively small but consistent improvements when topic models are used inside SMT models.

This work is different from previous works in the following aspects: 
\begin{itemize}
\item We use a bidirectional LSTM as an alternative to a feed-forward NN~\citep{Devlin:acl14} or maximum entropy-like classifiers~\citep{Tamchyna:acl6}.
\item In comprehensive experiments, we explore different methods to use additional meta-information in translation process.
\item We conduct a case study on e-commerce data where meta-information and context seem to be more effective. 
\item We perform human evaluation to confirm and explain improvements of automatic MT quality measures.
\end{itemize}

In our evaluation on an e-commerce translation test set containing a mix of product categories, we observed moderate improvements using different approaches introduced in the literature. This is in agreement with previous works. For specific product categories, however, we obtained large and significant improvements with each method. These experiments confirm the benefit of using larger context and meta-information in  translation. In addition, we found that the problem is far from being solved by the current approaches.

\section{MT Models Leveraging Meta-information}\label{sec:models}

\subsection{Sparse Lexical Features}

The main idea is to bias a SMT system towards the vocabulary and style of the target domain that can be inferred from the latent topics of the source sentence~\citep{ACL-2012-Eidelman}. We employ sparse lexical features~\citep{Hasler:iwslt12} and sparse topic features on top of common dense features in a SMT system. 
Sparse lexical features are tuples composed of a single source word and a single target word. These features can be also extended with another binary feature representing coexistence of a specific topic or text category in the source sentence. Topic information can be obtained from topic models trained on the source side of the bilingual training corpus along with other available in-domain monolingual data in the same language. The features with topic information are triggered by the topic of the source sentence, that is, for a particular source sentence to be translated, only the features that have been seen with the topic of that sentence will fire. 

We can also add information like topics or text categories for each phrase pair in the phrase-table. This information can be inferred from each phrase pair independent of the context or it can be inferred from the sentence pairs from which a given phrase pair is extracted. Therefore, each phrase pair is augmented with its topics, i.e., a vector of membership values of the phrase pair to each topic. The topic model is trained on an appropriate monolingual data in the source language. Then, based on the source side of each phrase or sentence pair, the topics and their probabilities, which again form a vector, are inferred from the topic model. We can combine both types of features to create a SMT more sensitive to the context, similar to~\citep{Mathur:mtsummit2015}.
The idea of a topic vector can be extended to any other context vector, i.e., the vector is simply a phrase-pair-specific representation of the meta-information in a continuous vector space. 

\subsection{Feed-Forward neural translation model}\label{sec:ffnm}
A neural network model previously used as a feature in a PBMT system is the neural network joint model (NNJM),
a feed forward architecture 
presented in~\citep{Devlin:acl14}. Assuming  
the target sentence $E:e_1,\dots,e_I$ and given the source sentence $F:f_1,\dots,f_J$, NNJM predicts a target word $e_i$ given a window $f_{b_i - w/2}^{b_i + w/2}$ of size $w+1$ around the corresponding source word and a target history $e_{i-v}^{i-1}$ of length $v$. 
This context is represented in the model by the concatenation of word embeddings corresponding to the source and target words. 
Because of the dependence on target context, the model has to be evaluated during search for each translation hypothesis.
Using a full softmax as output layer, this would be computationally prohibitive. Instead we use noise-contrastive estimation (NCE) as described in~\citep{Zoph:naacl16}. We rely on the self-normalizing property of NCE and do not perform a manual normalization of the network outputs. 
To further reduce evaluation time, the hidden layer contributions of all words in the vocabulary at all possible input positions are precomputed before search. 
This leads to a model which is fast enough to be evaluated during the search.

\begin{figure}[ht]
\begin{center}
\includegraphics[width=0.7\textwidth]{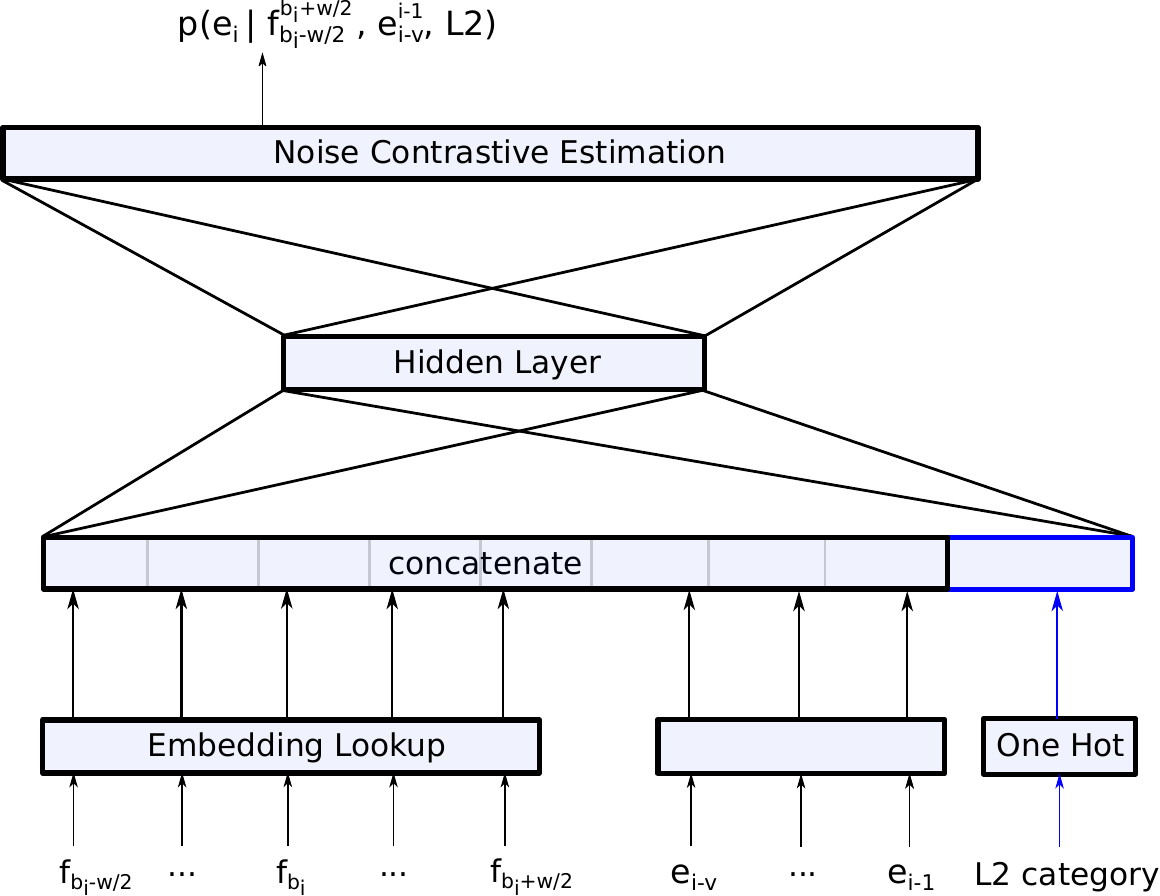}
\end{center}
\caption{\label{fig:nnjm-1hot} General architecture of neural network joint lexical model. 
Here, the $L_2$ product category is also shown as an optional input to the model.}
\end{figure}

Due to the their flexibility, neural networks models can be easily augmented with additional inputs to integrate any kind of context information.
In this work, we integrate the product category information into the feed-forward model by creating a one-hot category vector and appending it to the concatenation of the word embeddings, as shown in Figure~\ref{fig:nnjm-1hot}. The additional input can be included in the precomputation of the hidden layer in a straightforward manner, because it only adds one more input that can be looked up just like the ones from the source words. Variants like using an embedding for the category or appending it to the hidden layer have also been investigated. However, we have not seen significant differences in terms of the MT quality. 

\subsection{Bidirectional RNN translation model}\label{sec:btm}
Recurrent neural network models can be efficiently integrated into the framework of a phrase-based machine translation system under some circumstances~\citep{Alkhouli:wmt2015}. In this work, we have adopted the encoder part of the bidirectional encoder-decoder machine translation architecture described in~\citep{bahdanau2014neural}. This bidirectional translation model (BTM) takes the whole source sentence as input and estimates the probability $p(e_i|f_1^I, b_i)$ of a translation of the source word at position $b_i$. This architecture is also similar to~\citep{Sundermeyer:emnlp14}, but instead of the class-factored output layer we chose importance sampling~\citep{Jean:acl2015} to train the model. Also, instead of introducing $\epsilon$-tokens for unaligned words we obtain a unique $b_i$ by applying Devlin's affiliation heuristic to the statistical word alignments~\citep{Devlin:acl14}. Since the model does not depend on the target history the outputs for all combinations of $e_i$ and $b_i$ can be precomputed before the search. Thus, it is feasible to calculate the full softmax layer without approximations during the decoding.

\begin{figure}[ht]
\begin{center}
\includegraphics[width=0.75\textwidth]{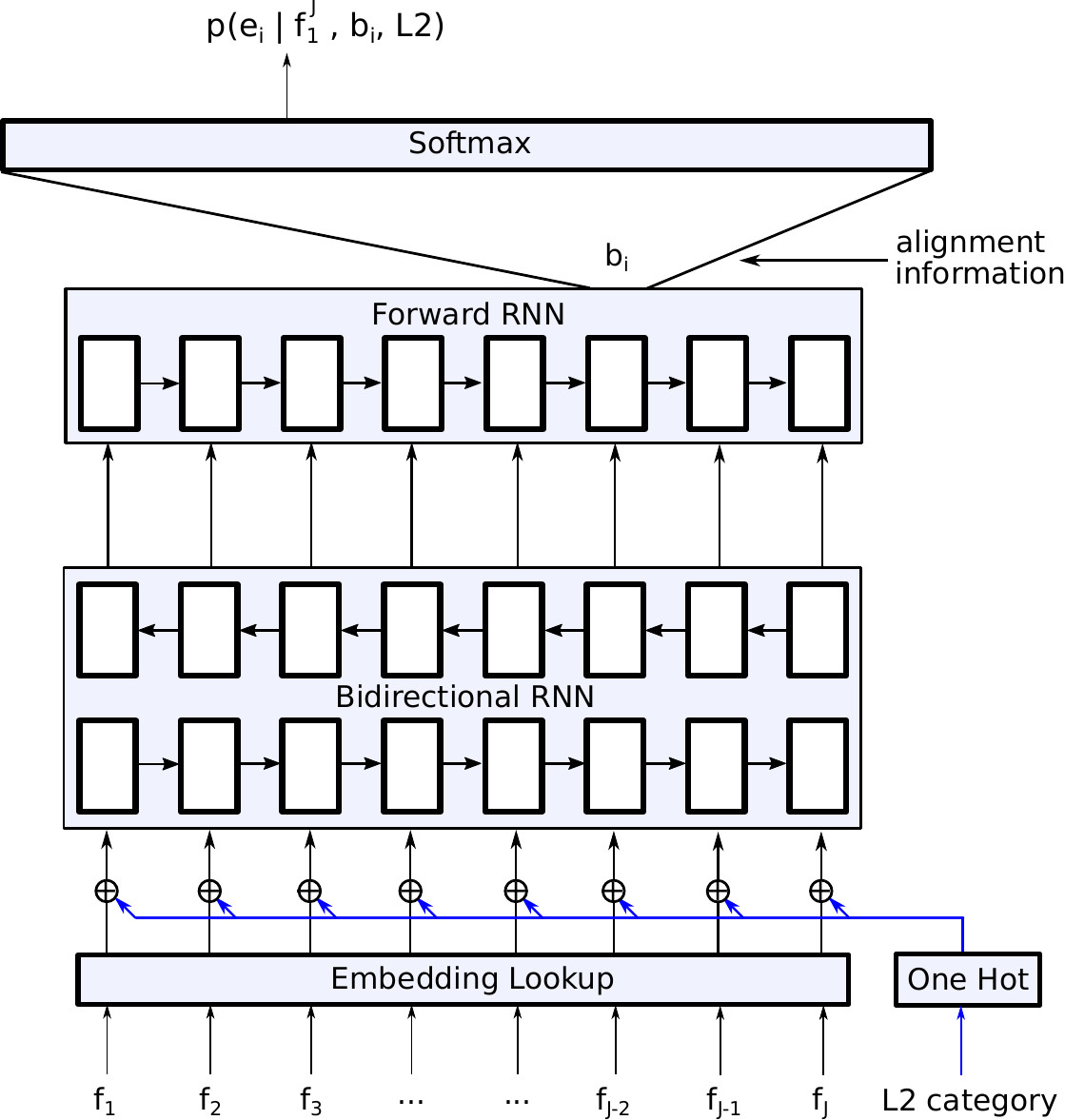}
\end{center}
\caption{\label{fig:btm-1hot} General architecture of bidirectional recurrent neural network lexical model. 
$L_2$ category is also shown as an optional input to the model.}
\end{figure}

Similar to the feed-forward model, we augment the BTM by adding category information as an additional input. In Figure~\ref{fig:btm-1hot}, an example of the resulting network architecture is shown. 
The one-hot category vector is concatenated to the word embedding at each source position. Other methods like treating the category as a special word in front of the actual sentence or replacing the one-hot vector with a category embedding have also been investigated, but their performance was worse or the same as with the architecture in Figure~\ref{fig:btm-1hot}.

\subsection{Category-aware generative model}\label{sec:genmodel}
Here, the main idea is to introduce category $L_2$ as a hidden variable into the translation probability $p(e|f)$ of the target word $e$ given the source word $f$ originating from a title of category $L_2$:
\begin{eqnarray}
\nonumber      p(e|f) & = & \sum_{L_2} p(e,L_2|f) \\ 
      			      & = & \sum_{L_2}p(e|f) \cdot p(L_2|e,f) \\
\nonumber             & = & \sum_{L_2} p(e|f) \cdot p(L_2|e)
\end{eqnarray}
where $p(L_2|e)$ is the probability of a category $L_2$ given a target word $e$. This probabilistic function implicitly penalizes word and phrase translations that are rarely observed in $L_2$, and it favors target words frequently appearing in $L_2$, but not in other categories. We should note that $p(L_2|e)$ is different from category-specific language models of type $p(e|L_2)$. The main advantage of this model is the possibility to be trained on larger amounts of additional in-domain monolingual data in the target language that has category information.

\section{Experiments}\label{sec:experiments}

We conduct comprehensive experiments with various methods to see the impact of meta information on the translation of item titles in the e-commerce domain. We use two baseline phrase-based MT systems. The first one is based on the Moses toolkit~\citep{MOSES-2007}, and the second one is an in-house phrase-decoder~\citep{iwslt10:EC:apptek}, which is similar to the Moses decoder. In both systems, we use standard SMT features, including word-level and phrase-level translation probabilities, the distortion model, and an n-gram LM. Due to the nature of the item titles,
we did not use any lexicalized reordering models in the MT system. The distortion limit was
set to 6. On the target side, we built a trigram LM, which is optimum on this task, using KenLM~\citep{KENLM-2011} trained with modified Kneser-Ney smoothing~\citep{Chen:1996}.
The LM is trained on the target side of bilingual data plus additional in-domain monolingual data composed of 60M words of item titles data.
In addition, we have also used a 5-gram operation sequence model (OSM)~\citep{OSM-2011-Durrani} and a 7-gram joint translation and reordering (JTR) model~\citep{Guta:JTR:wmt15}, which share the same idea and concept, in Moses-based and in-house systems, respectively. 
The feature weights are optimized using the k-best batch MIRA implementation provided in the Moses toolkit~\citep{Cherry:naacl12}. 
In the in-house decoder, the feature weights are tuned with minimum error rate training (MERT)~\cite{och2003mert} on $n$-best lists of the development set.
The MT quality is judged using the automatic case-insensitive 
BLEU~\citep{BLEU-2002-Papineni} and TER~\citep{TER-2006-Snover} scores. 
Statistical significance tests were conducted using approximate randomization tests~\citep{MultEval-2011-Clark}.

We have implemented the NNJM described in Section~\ref{sec:ffnm} and the BTM described in Section~\ref{sec:btm} using TensorFlow\footnote{www.tensorflow.org} 
The trained models are then exported to our in-house decoder using TensorFlow's C++ API. As the model is independent of the target history, we are able to precalculate scores for all phrase pairs that are selected in the phrase matching step for a given source sentence. This improves the runtime cost, as the model does not need to be queried during the decoding.

For training and evaluating feed-forward neural network models in Moses, we rely upon the Neural Probabilistic Language Model Toolkit (NPLM) \citep{vaswani-EtAl:2013:EMNLP}. NPLM can be used to train both neural language models and joint models. We also integrate the models into our in-house decoder as language models.

\begin{table}[ht]
\begin{center}
\begin{tabular}{|lr|c|c|} \hline
       &              &  English     &   Italian      \\ \hline \hline
Train: & Sentences    & \multicolumn{2}{|c|}{10,231,392}\\ \hline
       & Tokens       & 117 M & 115 M \\ \hline
       & Vocabulary   & 493 K & 582 K\\ \hline
       & Singletons   & 239 K & 257 K\\ \hline
 \hline
Dev.:  & Sentences    & 910 & 910\\ \hline
       & Tokens       & 10,818 & 11,159\\ \hline
       & Vocabulary   & 4,422 & 4,481\\ \hline
       & OOVs         & 321 & 310\\ \hline
 \hline
Test   & Sentences    & 910 & 910\\ \hline
       & Tokens       & 10,814 & 11,241\\ \hline
       & Vocabulary   & 4,487 & 4,532\\ \hline
       & OOVs         & 337 & 321 \\ \hline
\end{tabular}
\caption{ Corpus Statistics }\label{corp:stats}
\end{center}
\end{table}

We conduct experiments on an English-to-Italian e-commerce item titles translation task. The training data is composed of in-house data (item titles, descriptions, etc.) as in-domain corpus and also publicly available data that has been sampled according to the similarity to in-domain data. The corpus statistics are summarized in Table~\ref{corp:stats}. The size of in-domain data is about 4.6\% of the training data in terms of source side tokens. Before calculating the corpus statistics, we apply some usual text pre-processing including tokenization and replacement of numbers with a placeholder token; and also some domain dependent processing such as replacing product specification - e.g., \texttt{6S}, and \texttt{1080p} -  with a general token. 

The in-domain data has also some meta information that identifies the category of each sentence/segment.
We denote this meta-information as $L_2$. 
In all experiments we define six categories, five selected $L_2$ categories plus one {\em other} category.
Meta information for out-domain training data is inferred based on a state-of-the-art text classification algorithm trained on a big in-domain source monolingual data, for which the $L_2$ category is available. The Dev and Test sets also contain the category information, and they have two human reference translations. 

\begin{table}[ht]
\begin{center}
\begin{tabular}{|l|l||r|r|} \hline 
    &  & \multicolumn{2}{|c|}{\bf Test} \\ \cline{3-4}
\# & \multicolumn{1}{|c||}{\bf System} & \bf BLEU & \bf TER \\ 
    &                                   & \small\bf [\%] & \small\bf [\%] \\ \hline \hline
1 & Moses Baseline & 37.4 & 45.7\\ \hline
2 & + \citet{Tamchyna:acl6} & 37.4 & 45.9 \\ \hline
3 & + Word-pair SF & 37.6 & 45.9 \\ \hline
4 & + Category SF & 38.3 & 44.8 \\ \hline
5 & + Category SF +\citet{Mathur:mtsummit2015} & 37.5 & 45.4 \\ \hline
6 & + NNJM & 37.7  & 45.0\\ \hline
\hline
7 & In-house Baseline & 37.0 & 46.0 \\ \hline
8 &  + BTM & 37.7 & 45.5 \\ \hline
9 &  + NNJM & 37.5 & 45.5 \\ \hline \hline
10 & Pure NMT baseline & 28.0 & 54.9 \\ \hline
\end{tabular}
\end{center}
\caption{\label{table1}  Experimental results: English-to-Italian item title translation task.}
\end{table}

In Table~\ref{table1}, the translation results are shown in terms of both BLEU score and TER. Moses baseline has a slightly better quality performance compared to in-house baseline system, seventh row. We report in-house results since NNJM and BTM models described respectively in Section~\ref{sec:ffnm} and Section~\ref{sec:btm}, are implemented in this in-house system. 
We also report the results of a state-of-the-art baseline NMT system as described in \citep{chen2016guided}. Based on  \citet{chen2016guided} and also our experiments on another language pair in the same task, there is room to improve this baseline using techniques like back-translation~\citep{sennrich2015improving} and guided-alignment, but the performance of a single system (and even with ensembling technique) is lower than a strong phrase-based baseline. The lower performance is due to the nature of item titles data that are not appropriate for NMT approach. Item titles data are not grammatical,  they are very irregular and like other user-generated data very noisy. Therefore, we confine to report a simple NMT baseline to show the characteristics of the task. 

System in the second row is based on the work of~\citet{Tamchyna:acl6}, we have used the default features proposed in the original work: source bag of words, target bi-grams, source indicator and target indicator. We have also conducted two experiments to train the classifier model on in-domain data and on mixed domain data, we report the translation results if the classifier model is trained only on in-domain data. 
In third and fourth experiments in Table~\ref{table1}, we use word-pair feature, and word-pair plus category information as sparse features (SF), respectively. We adopt the method presented in \citep{Hasler:iwslt12} in our case study, we replace topic models with predefined category information.
We include the work of \citet{Mathur:mtsummit2015} in our experiments, with this difference that we use category information instead of topic models. As required by this model, we have augmented to each phrase of the phrase table a six-value normalized vector that represents the membership value of each phrase to each category. These membership values are simply normalized frequencies of categories in the parent sentence pairs of a phrase pair. Parent sentence pairs are those that include a given phrase pair. 
The next systems in the table are based on neural network models. In NNJM, we set a window of four source words. Since the translation of polysemous words is not directly dependent on the previous target words, and also, to make this model more comparable to BTM,  we do not use target words in NNJM.
In NNJM, input word embedding and output word embedding are set to 150 and 750, respectively. We use a single hidden layer in NNJM. Despite the same setting in sixth and ninth rows, the implementation of NNJM in the sixth row is based on NPLM toolkit in Moses, and implementation in ninth row is as described in Section~\ref{sec:ffnm} in TensorFlow toolkit.
In BTM, we have used the embedding size of 620, RNN size of 1000 and GRU cells. The learning rate was set to 0.0002, decaying by 0.9 each epoch. 
The vocabulary is set to most frequent 100,000 words for both NNJM and BTM.

\begin{table*}
\begin{center}
\resizebox{\columnwidth}{!}{
\begin{tabular}{|l||r|r||r|r||r|r||r|r||r|r|} \hline 
 & \multicolumn{2}{|c||}{Cat. I} & \multicolumn{2}{|c||}{Cat. II} & \multicolumn{2}{|c||}{Cat. III} & \multicolumn{2}{|c||}{Cat. IV} & \multicolumn{2}{|c|}{Cat. V}  \\ \cline{2-11}
\multicolumn{1}{|c||}{Sys.} & {\small BLEU} & {\small TER} & {\small BLEU} & {\small TER} & {\small BLEU} & {\small TER} & {\small BLEU} & {\small TER} & {\small BLEU} & {\small TER} \\ \hline \hline

Moses Baseline & 58.5  & 27.3  & 36.2  & 44.2  & 35.8  & 47.2  & 29.6  & 51.7  & 34.1  & 48.2  \\ \hline
 + Tamchyna et al. 2016 & 58.5  & 29.6  & 37.4  & 45.5  & 35.0  & 48.5  & 30.0  & 51.9  & 34.0  & 49.0   \\ \hline
 + Word-pair SF & 58.6  & 28.7  & 35.7  & 45.3  & 36.0  & 47.7  & 29.3  & 51.7  & 33.7  & 49.5   \\ \hline
 + Category SF & 60.6  & 27.1  & 36.2  & 43.9  & 35.8  & 47.7  & 30.1  & 50.2  & 34.0  & 48.7   \\ \hline
 + Category SF  & 59.3  & 28.0  & 36.8  & 42.8  & 38.0  & 46.9  & 29.6  & 50.7  & 34.3  & 48.2   \\ 
 + Mathur et al. 2015 & & & & & & & & & & \\ \hline
 + NNJM & 58.5  & 28.7  & 36.2  & 44.3  & 36.0  & 46.9  & 33.6  & 48.6  & 33.8  & 47.1  \\  \hline \hline 
In-house Baseline & 57.7  & 28.5  & 36.9  & 44.3  & 34.9  & 48.0  & 29.2  & 52.4  & 33.9  & 47.7  \\ \hline
 + BTM & 59.7  & 27.8  & 37.7  & 44.1  & 36.4  & 46.7  & 28.1  & 51.7  & 34.4  & 46.6  \\ \hline
 + NNJM & 59.9  & 27.8  & 37.8  & 43.4  & 36.6  & 46.4  & 32.2  & 50.0  & 35.1  & 46.6  \\ \hline \hline
 \# test sentences & \multicolumn{2}{|c||}{33} & \multicolumn{2}{|c||}{62} & \multicolumn{2}{|c||}{27} & \multicolumn{2}{|c||}{32} & \multicolumn{2}{|c|}{30} \\ \hline

\end{tabular}
}
\end{center}
\caption{Translation of English titles from five selected product categories into Italian (BLEU scores and TER in \%).}
\label{table3}
\end{table*}

As shown in Table~\ref{table1}, the differences of MT systems are relatively small. Among other reasons, limited number of samples per each targeted categories in the test set could explain these small differences. The number of occurrences for each category are shown in the last row of Table~\ref{table3}, a similar distribution of categories exists for the development set.

\begin{figure}[ht]
\begin{center}

\begin{minipage}[t]{\linewidth}
\includegraphics[width=\textwidth]{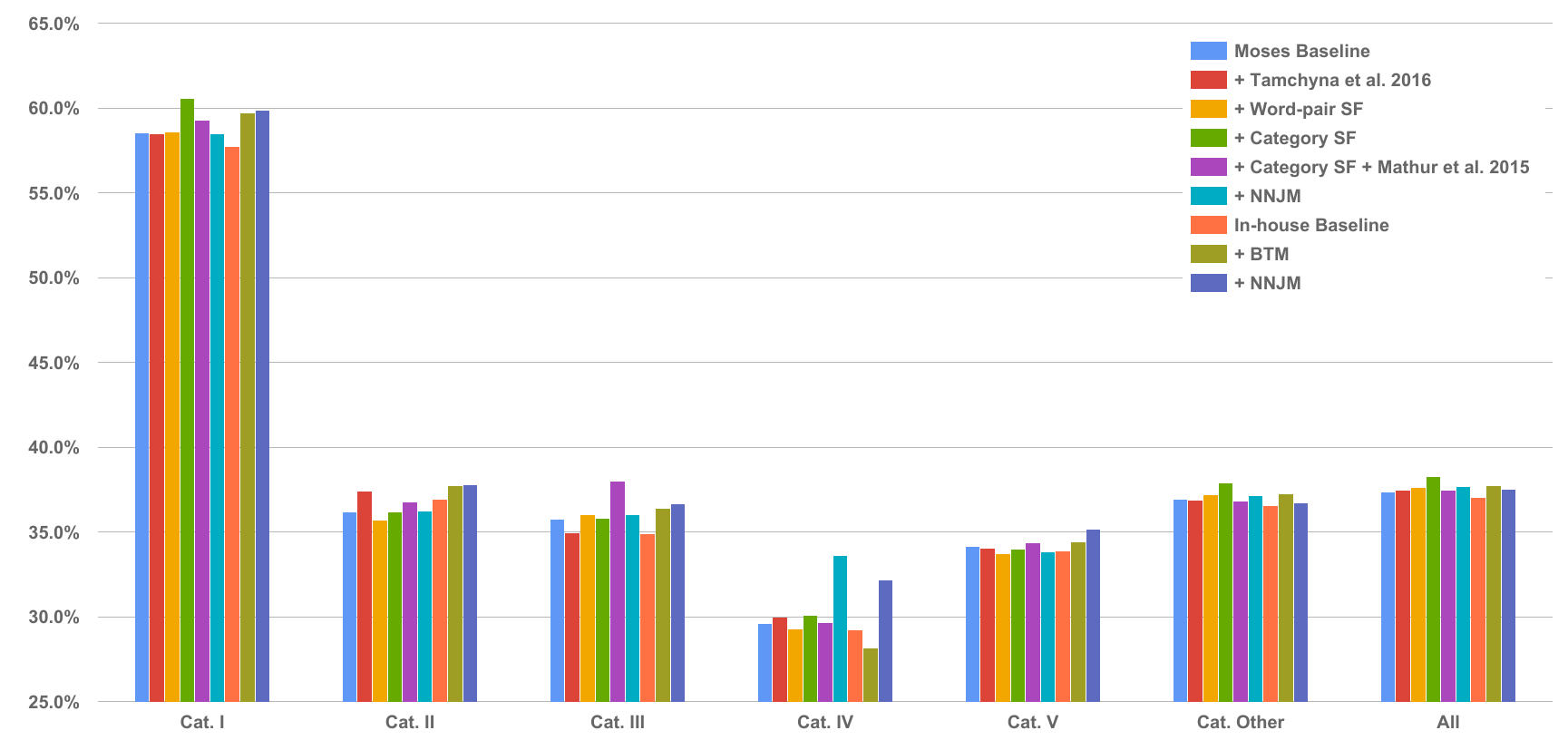}
\caption{\label{fig:cat-bleu-res} Detailed BLEU results.}
\end{minipage}

\begin{minipage}[t]{\linewidth}
\includegraphics[width=\textwidth]{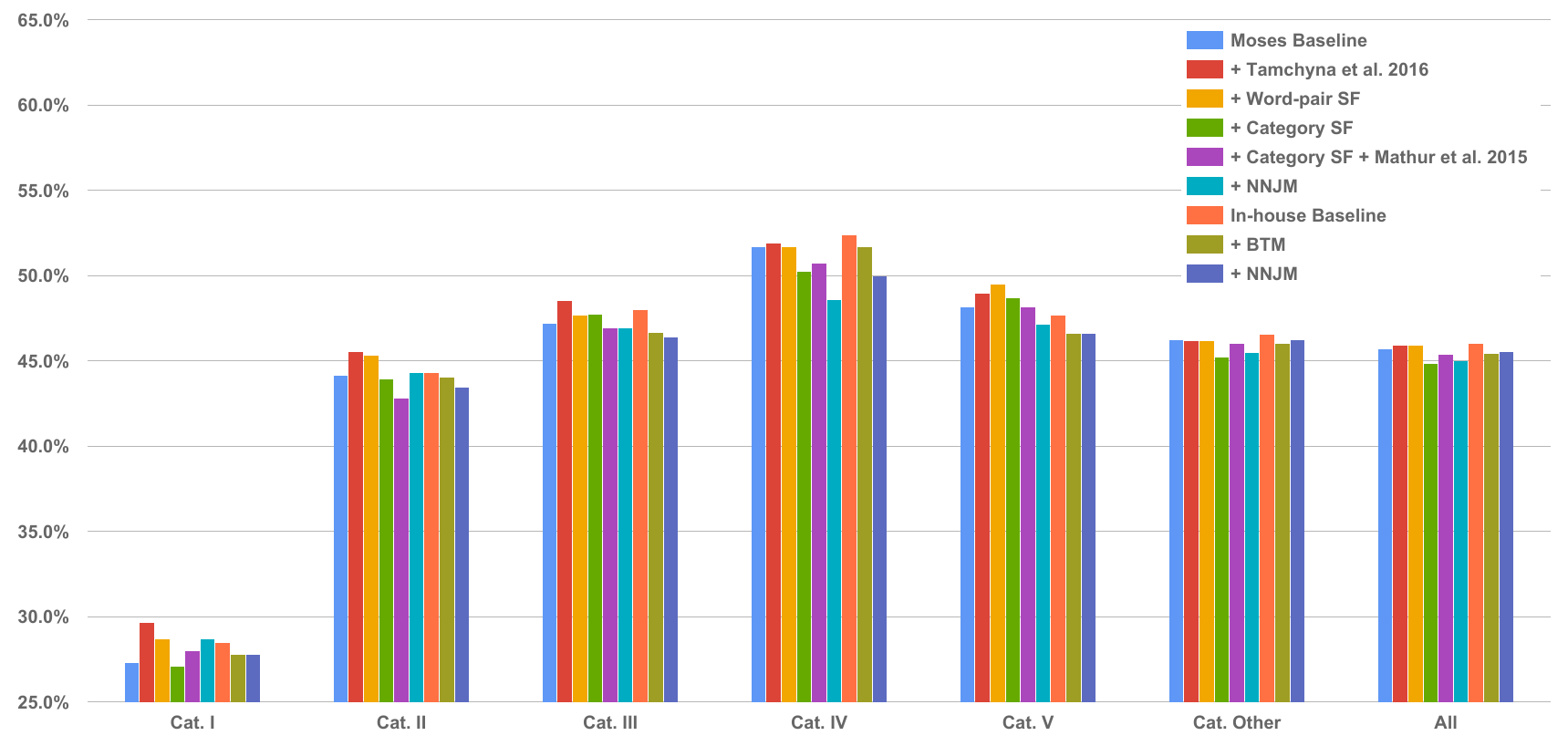}
\caption{\label{fig:cat-ter-res} Detailed TER results.}
\end{minipage}

\end{center}

\end{figure}

In Table~\ref{table3}, we have also shown the detailed improvements achieved on five specific $L_2$ categories. 
%
Now, we observe much larger improvements in the selected categories. We observe the best performing system in Table~\ref{table1} is not necessarily the best system for our specific goal that we embed corresponding additional information into the translation process. As shown in Table~\ref{table3}, in-house system plus NNJM has the most consistent improvements over its corresponding baseline.
For a better comparison of results, you may use diagrams in Figure~\ref{fig:cat-bleu-res} and Figure~\ref{fig:cat-ter-res}.
Please note that the results for all neural models are without category information so far. Although we expect improvements of MT quality on the particular categories by adding the category information as an additional signal into the neural models, we have not seen any significant improvements. We thought, we may need to send a stronger signal in the training process and therefore we have tried different ways to embed category information into the model, but the results in all cases were almost at the same level. These results are in contrast with the results reported in~\citep{chen2016guided}, they have shown some improvements using category information in an NMT architecture. This disagreement might be due to different architectures and also due to different baselines. 

To investigate why category information does not contribute to MT quality improvements in our experiments, we have conducted some experiments to measure the perplexity on the training corpus when the BTM model is trained with and without $L_2$ information and we have observed a small increase in perplexity when we used the $L_2$ categories. We may discuss these results in two directions, first the category information are too sparse in our settings to be useful, and second, the category information has no more information over the text itself, especially when the text is processed globally in a neural network model. 
%
%

We have also conducted an in-house human analysis of polysemous words to better understand the situation. We observe some examples where $L_2$ categories help to disambiguate the meaning of the words. At the same time, there are other polysemous words for which category information cannot help. For example, the sense of the word \texttt{Vans} can be identified if we know it is from \texttt{Motors} category meaning plural of \texttt{Van} or from \texttt{Clothing} category meaning a brand name. Another example is word \texttt{mixer} that has two different meaning in \texttt{Kitchen} and in \texttt{Music instruments} categories. However, there are cases that product categories will not help to disambiguate the meaning of the word, e.g., word \texttt{Ship} in category \texttt{Toys} may have at least two different meanings as a noun or as a verb.  
Therefore, $L_1$ or $L_2$ categories information might be not helpful for some polysemous words. 


The system described in Section~\ref{sec:genmodel} uses an additional probability $p(L|e)$ for  the input sentence category given a target word candidate. We have implemented it as an additional lexical model that assigned a score to each target phrase pair. Using this model with a tuned weight in the log-linear model combination did not result in significant improvement of automatic MT measures, the BLEU/TER scores remained basically the same as in-house baseline, line 7 of Table~\ref{table1}. However, this system was stable in the sense that its translations did not differ much from the baseline, and the observed changes predominantly affected content words. A manual analysis showed several examples where translations of polysemous words which were inappropriate for a certain product category were changed for the better.  For instance, the Italian translation of the term \texttt{latte} used in an English title in the category \texttt{Bags and Accessories} as a bag's color was corrected from \texttt{latte macchiato} to \texttt{color crema}. Also, in the \texttt{Kitchen Appliances} category, the word \texttt{tamper} was incorrectly translated as  \texttt{manomissione (fabrication)} by the baseline system.  The system that uses the category-specific word lexicon $p(L|e)$ has correctly translated it as \texttt{pestello}.

In general, systems with the category-aware features, implicitly or explicitly, cause moderate improvements compared to a baseline system in terms of automatic MT error measures like BLEU. 
However, it can be argued that these automatic MT measures do not reflect the impact of polysemous words, since such words occur rarely as compared to all other words even if we consider all polysemous words, not only those which have a wrong translation in the baseline. 
A human evaluation focusing on translation of polysemous words can potentially justify the soundness of the employed method. 
For such an evaluation, we randomly selected 300 item titles if they had at least one of the polysemous words which have been identified as such by domain expert linguists. We asked human translators to answer five questions for each translation, the most important and relevant question was whether the identified polysemous word in the text was correctly translated or not. 
The human evaluation results show that although the proposed system cause moderate improvements - about the same as we observe in the above tables - one third of the polysemous words have not been translated correctly. 

We attribute the weak performance of the presented models on polysemous words to the bias that we observe in the training data to the most frequent meanings of such words. For example, the translation of \texttt{Apple} as a brand name is by far more frequent in our training data than translation of this word as a common noun with the meaning of a fruit.
Optimization for the BLEU score seems to additionally increase the bias, since the tuning set is in most cases also biased to the most frequent meaning of such words.

\section{Conclusion}\label{sec:conclusion}
We employed several different ways to incorporate explicit meta-information or larger context to better translate polysemous words or improve MT quality in general. We explored existing state-of-the-art methods  that can potentially help in this task or can accept another input as meta-information, e.g., \citep{Hasler:iwslt12,Mathur:mtsummit2015,Tamchyna:acl6,Devlin:acl14}. 

To better exploit the source-side topic/category labels, we introduced a bidirectional LSTM to encode the entire source sentence context to translate a word. We investigated different ways of incorporating meta-information in the encoder. In addition, we proposed a novel generative model that can leverage topic-labeled target monolingual data.

We conducted comprehensive experiments on different ways of using additional meta-information in translation process, including both the given human-labeled and the automatically predicted meta-information. Our case study was an e-commerce English-to-Italian translation task.
We observed improvements up to 3\% in terms of the BLEU score for some input text categories. Finally, we performed a human evaluation to confirm and explain improvements of automatic MT quality measures. We have realized that, although the observed improvements were reconfirmed by human evaluation, there were many polysemous words in the test set that were still not translated correctly. The take-home message of this research is that the problem of how to best use meta-information in MT for correct, in-topic translation of polysemous words and phrases is still far from being solved.  

In the future, we aim to use other types of meta-information that may be better suited for disambiguating the meaning of polysemous words. For example, we plan to leverage automatically predicted domain-specific named-entity tags as meta-data for translation. Another area for future work is how to use meta-information more effectively, overcoming data sparseness and bias problems. 

In the future, we also plan to adopt a hybrid NMT and SMT approach similar to ~\citep{leonard:hybrid_nmt_smt} to improve the translation of polysemous words in item title domain. In this way, we can benefit from long context coverage of NMT system to find a more appropriate translation based on the context for polysemous words, and also benefit from more control over the generation process of SMT system and its features like text override.

\bibliographystyle{apalike}

\bibliography{mtsummit2017}

\begin{thebibliography}{}

\bibitem[Alkhouli et~al., 2015]{Alkhouli:wmt2015}
Alkhouli, T., Rietig, F., and Ney, H. (2015).
\newblock Investigations on phrase-based decoding with recurrent neural network
  language and translation models.
\newblock In {\em Proceedings of the Tenth Workshop on Statistical Machine
  Translation, WMT@EMNLP 2015, 17-18 September 2015, Lisbon, Portugal}, pages
  294--303.

\bibitem[Bahdanau et~al., 2014]{bahdanau2014neural}
Bahdanau, D., Cho, K., and Bengio, Y. (2014).
\newblock Neural machine translation by jointly learning to align and
  translate.

\bibitem[Carpuat and Wu, 2007]{Carpuat:emnlp2007}
Carpuat, M. and Wu, D. (2007).
\newblock Improving statistical machine translation using word sense
  disambiguation.
\newblock In {\em EMNLP-CoNLL 2007, Proceedings of the 2007 Joint Conference on
  Empirical Methods in Natural Language Processing and Computational Natural
  Language Learning, June 28-30, 2007, Prague, Czech Republic}, pages 61--72.

\bibitem[Chen and Goodman, 1996]{Chen:1996}
Chen, S.~F. and Goodman, J. (1996).
\newblock An empirical study of smoothing techniques for language modeling.
\newblock In {\em Proceedings of the 34th Annual Meeting on Association for
  Computational Linguistics}, ACL '96, pages 310--318, Stroudsburg, PA, USA.
  Association for Computational Linguistics.

\bibitem[Chen et~al., 2016]{chen2016guided}
Chen, W., Matusov, E., Khadivi, S., and Peter, J.-T. (2016).
\newblock Guided alignment training for topic-aware neural machine translation.
\newblock {\em AMTA 2016, Vol.}, page 121.

\bibitem[Cherry and Foster, 2012]{Cherry:naacl12}
Cherry, C. and Foster, G.~F. (2012).
\newblock Batch tuning strategies for statistical machine translation.
\newblock In {\em Human Language Technologies: Conference of the North American
  Chapter of the Association of Computational Linguistics, Proceedings, June
  3-8, 2012, Montr{\'{e}}al, Canada}, pages 427--436.

\bibitem[Clark et~al., 2011]{MultEval-2011-Clark}
Clark, J.~H., Dyer, C., Lavie, A., and Smith, N.~A. (2011).
\newblock Better hypothesis testing for statistical machine translation:
  Controlling for optimizer instability.
\newblock In {\em Proceedings of the 49th Annual Meeting of the Association for
  Computational Linguistics: Human Language Technologies: short papers-Volume
  2}, pages 176--181. Association for Computational Linguistics.

\bibitem[Dahlmann et~al., 2017]{leonard:hybrid_nmt_smt}
Dahlmann, L., Matusov, E., Petrushkov, P., and Khadivi, S. (2017).
\newblock Neural machine translation leveraging phrase-based models in a hybrid
  search.
\newblock In {\em Proceedings of the Empirical Methods in Natural Language
  Processing (EMNLP)}.

\bibitem[Devlin et~al., 2014]{Devlin:acl14}
Devlin, J., Zbib, R., Huang, Z., Lamar, T., Schwartz, R.~M., and Makhoul, J.
  (2014).
\newblock Fast and robust neural network joint models for statistical machine
  translation.
\newblock In {\em Proceedings of the 52nd Annual Meeting of the Association for
  Computational Linguistics, {ACL} 2014, June 22-27, 2014, Baltimore, MD, USA,
  Volume 1: Long Papers}, pages 1370--1380.

\bibitem[Durrani et~al., 2011]{OSM-2011-Durrani}
Durrani, N., Schmid, H., and Fraser, A.~M. (2011).
\newblock A joint sequence translation model with integrated reordering.
\newblock In {\em The 49th Annual Meeting of the Association for Computational
  Linguistics: Human Language Technologies, Proceedings of the Conference,
  19-24 June, 2011, Portland, Oregon, {USA}}, pages 1045--1054.

\bibitem[Eidelman et~al., 2012]{ACL-2012-Eidelman}
Eidelman, V., Boyd-Graber, J., and Resnik, P. (2012).
\newblock {Topic Models for Dynamic Translation Model Adaptation}.
\newblock In {\em Proceedings of the 50th Annual Meeting of the Association for
  Computational Linguistics (Volume 2: Short Papers)}, pages 115--119, Jeju
  Island, Korea. Association for Computational Linguistics.

\bibitem[Guta et~al., 2015]{Guta:JTR:wmt15}
Guta, A., Wuebker, J., Graca, M., Kim, Y., and Ney, H. (2015).
\newblock Extended translation models in phrase-based decoding.
\newblock In {\em Proceedings of the Tenth Workshop on Statistical Machine
  Translation, WMT@EMNLP 2015, 17-18 September 2015, Lisbon, Portugal}, pages
  282--293.

\bibitem[Hasler et~al., 2014a]{EACL-2014-Hasler}
Hasler, E., Blunsom, P., Koehn, P., and Haddow, B. (2014a).
\newblock {Dynamic Topic Adaptation for Phrase-based MT}.
\newblock In {\em Proceedings of the 14th Conference of the European Chapter of
  the Association for Computational Linguistics}, pages 328--337, Gothenburg,
  Sweden. Association for Computational Linguistics.

\bibitem[Hasler et~al., 2012a]{IWSLT-2012-Hasler}
Hasler, E., Haddow, B., and Koehn, P. (2012a).
\newblock {Sparse Lexicalised Features and Topic Adaptation for SMT}.
\newblock In {\em Proceedings of the Workshop on Spoken Language Translation
  (IWSLT)}, pages xxx--xxx.

\bibitem[Hasler et~al., 2012b]{Hasler:iwslt12}
Hasler, E., Haddow, B., and Koehn, P. (2012b).
\newblock Sparse lexicalised features and topic adaptation for {SMT}.
\newblock In {\em International Workshop on Spoken Language Translation,
  {IWSLT} 2012, Hong Kong, December 6-7, 2012}, pages 268--275.

\bibitem[Hasler et~al., 2014b]{AMTA-2014-Hasler}
Hasler, E., Haddow, B., and Koehn, P. (2014b).
\newblock {Combining domain and topic adaptation for SMT}.
\newblock In {\em Proceedings of the Eleventh Conference of the Association for
  Machine Translation in the Americas (AMTA)}, volume~1, pages 139--151.

\bibitem[Heafield, 2011]{KENLM-2011}
Heafield, K. (2011).
\newblock Kenlm: Faster and smaller language model queries.
\newblock In {\em Proceedings of the Sixth Workshop on Statistical Machine
  Translation}, pages 187--197. Association for Computational Linguistics.

\bibitem[Jean et~al., 2015]{Jean:acl2015}
Jean, S., Cho, K., Memisevic, R., and Bengio, Y. (2015).
\newblock On using very large target vocabulary for neural machine translation.
\newblock In {\em Proceedings of the 53rd Annual Meeting of the Association for
  Computational Linguistics and the 7th International Joint Conference on
  Natural Language Processing of the Asian Federation of Natural Language
  Processing, {ACL} 2015, July 26-31, 2015, Beijing, China, Volume 1: Long
  Papers}, pages 1--10.

\bibitem[Koehn et~al., 2007]{MOSES-2007}
Koehn, P., Hoang, H., Birch, A., Callison-Burch, C., Federico, M., Bertoldi,
  N., Cowan, B., Shen, W., Moran, C., Zens, R., et~al. (2007).
\newblock Moses: Open source toolkit for statistical machine translation.
\newblock In {\em Proceedings of the 45th annual meeting of the ACL on
  interactive poster and demonstration sessions}, pages 177--180. Association
  for Computational Linguistics.

\bibitem[Mathur et~al., 2015]{Mathur:mtsummit2015}
Mathur, P., Federico, M., K{\"o}pr{\"u}, S., Khadivi, S., and Sawaf, H. (2015).
\newblock Topic adaptation for machine translation of e-commerce content.
\newblock {\em Proceedings of MT Summit XV}, page 270.

\bibitem[Matusov and K\"{o}pr\"{u}, 2010]{iwslt10:EC:apptek}
Matusov, E. and K\"{o}pr\"{u}, S. (2010).
\newblock {AppTek's APT Machine Translation System for IWSLT 2010}.
\newblock In Federico, M., Lane, I., Paul, M., and Yvon, F., editors, {\em
  International Workshop on Spoken Language Translation, IWSLT}, pages 29--36.

\bibitem[Mauser et~al., 2009]{Mauser:emnlp09}
Mauser, A., Hasan, S., and Ney, H. (2009).
\newblock Extending statistical machine translation with discriminative and
  trigger-based lexicon models.
\newblock In {\em Proceedings of the 2009 Conference on Empirical Methods in
  Natural Language Processing, {EMNLP} 2009, 6-7 August 2009, Singapore, {A}
  meeting of SIGDAT, a Special Interest Group of the {ACL}}, pages 210--218.

\bibitem[Och, 2003]{och2003mert}
Och, F.~J. (2003).
\newblock Minimum error rate training in statistical machine translation.
\newblock In {\em Proceedings of the 41st Annual Meeting on Association for
  Computational Linguistics}, pages 160--167.

\bibitem[Papineni et~al., 2002]{BLEU-2002-Papineni}
Papineni, K., Roukos, S., Ward, T., and Zhu, W.-J. (2002).
\newblock Bleu: a method for automatic evaluation of machine translation.
\newblock In {\em Proceedings of the 40th annual meeting on association for
  computational linguistics}, pages 311--318. Association for Computational
  Linguistics.

\bibitem[Sennrich et~al., 2015]{sennrich2015improving}
Sennrich, R., Haddow, B., and Birch, A. (2015).
\newblock Improving neural machine translation models with monolingual data.
\newblock {\em arXiv preprint arXiv:1511.06709}.

\bibitem[Snover et~al., 2006]{TER-2006-Snover}
Snover, M., Dorr, B., Schwartz, R., Micciulla, L., and Makhoul, J. (2006).
\newblock A study of translation edit rate with targeted human annotation.
\newblock In {\em Proceedings of association for machine translation in the
  Americas}, pages 223--231.

\bibitem[Sundermeyer et~al., 2014]{Sundermeyer:emnlp14}
Sundermeyer, M., Alkhouli, T., Wuebker, J., and Ney, H. (2014).
\newblock Translation modeling with bidirectional recurrent neural networks.
\newblock In {\em Proceedings of the 2014 Conference on Empirical Methods in
  Natural Language Processing, {EMNLP} 2014, October 25-29, 2014, Doha, Qatar,
  {A} meeting of SIGDAT, a Special Interest Group of the {ACL}}, pages 14--25.

\bibitem[Tamchyna et~al., 2016]{Tamchyna:acl6}
Tamchyna, A., Fraser, A.~M., Bojar, O., and Junczys{-}Dowmunt, M. (2016).
\newblock Target-side context for discriminative models in statistical machine
  translation.
\newblock In {\em Proceedings of the 54th Annual Meeting of the Association for
  Computational Linguistics, {ACL} 2016, August 7-12, 2016, Berlin, Germany,
  Volume 1: Long Papers}.

\bibitem[Vaswani et~al., 2013]{vaswani-EtAl:2013:EMNLP}
Vaswani, A., Zhao, Y., Fossum, V., and Chiang, D. (2013).
\newblock Decoding with large-scale neural language models improves
  translation.
\newblock In {\em Proceedings of the 2013 Conference on Empirical Methods in
  Natural Language Processing}, pages 1387--1392, Seattle, Washington, USA.
  Association for Computational Linguistics.

\bibitem[Zoph et~al., 2016]{Zoph:naacl16}
Zoph, B., Vaswani, A., May, J., and Knight, K. (2016).
\newblock Simple, fast noise-contrastive estimation for large {RNN}
  vocabularies.
\newblock In {\em {NAACL} {HLT} 2016, The 2016 Conference of the North American
  Chapter of the Association for Computational Linguistics: Human Language
  Technologies, San Diego California, USA, June 12-17, 2016}, pages 1217--1222.

\end{thebibliography}




\end{document}